\documentclass{poster19}
\begin{document}

%
\title{Hybrid Code Networks using a convolutional neural network as an input layer achieves higher turn accuracy}
%
\headtitle{PETR MAREK, HYBRID CODE NETWORKS USING A CONVOLUTIONAL NEURAL NETWORK AS AN INPUT LAYER ACHIEVES HIGHER TURN ACCURACY}

%
\author{Petr Marek}
%
\affiliation{%
Dept. of Cybernetics, Czech Technical University, Technick\'a 2, 166 27 Praha, Czech Republic}
  \email{marekp17@fel.cvut.cz}

\maketitle


\begin{abstract}
The dialogue management is a task of conversational artificial intelligence. The goal of the dialogue manager is to select the appropriate response to the conversational partner conditioned by the input message and recent dialogue state. Hybrid Code Networks is one of the models of dialogue managers, which uses an average of word embeddings and bag-of-words as input features. We perform experiments on Dialogue bAbI Task 6 and Alquist Conversational Dataset. The experiments show that the convolutional neural network used as an input layer of the Hybrid Code Network improves the model's turn accuracy.

\end{abstract}



\section{Introduction}
There was a significant spread of personal assistants and chatbots in recent years. Users have greater demands on their capabilities, as they become more mainstream and popular. The demand drives the research of better technologies of conversational artificial intelligence. One of the technologies is dialogue managers. The role of the dialogue manager is to select the most appropriate response based on the recent user's message and state of the dialogue.

Hybrid Code Networks is a type of dialogue manager, which combines a recurrent neural network with the domain-specific rules. Rules allow the model to learn with a considerably reduced amount of training examples.

The baseline Hybrid Code Networks uses an average of word embeddings and bag-of-words of user's message as input features. These features do not capture the order of the words in the message. We propose two new architectures inspired by Hybrid Code Networks, which can capture the order of words. The first architecture uses a convolutional neural network and the second architecture uses a recurrent neural network as input layers.

\begin{figure}[]
\begin{center}
\includegraphics[width=\linewidth]{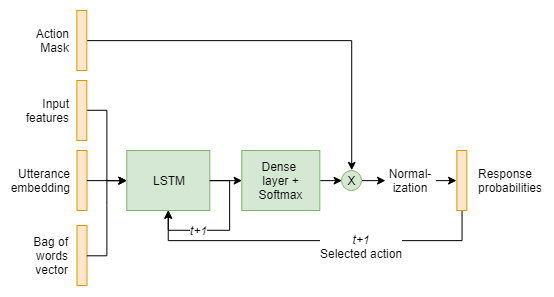}
\caption{Schema of baseline Hybrid Code Networks} 
\label{fig:SchemaOfHCN}
\end{center}
\end{figure}

\section{Related work}
There are various approaches to dialogue management. Dialogue managers can be divided into retrieval and generative dialogue managers, and into rule-based, end-to-end and hybrid-based dialogue managers.

The rule-based dialogue managers produce a response based on the set of rules \cite{bAbI}. The rules detect the presence of patterns in the user's input messages and the state of the dialogue. The rule-based systems work well in constrained domains, but they become more inaccurate and more challenging to implement as the complexity of the domain grows.

Dialogue managers working with text similarity measures the similarity between the user's input message and set of reference sentences. The most similar sentence out of the set of reference sentences determines the next action of the dialogue manager. The similarity is measured between vector representations of the message and sentences \cite{bAbI}. Vector  representations can be created by TF-IDF or average of word embeddings like word2vec \cite{Word2Vec}, GloVe \cite{Glove} or fastText \cite{fastText}. The similarity function can be cosine similarity or function learned by supervised learning \cite{SupervisedSemanticIndexing}.

Seq2seq dialogue managers learn a mapping between input message and output response \cite{seq2seq}. It consists of encoder and decoder recurrent neural networks. The encoder maps input message into a fixed size vector representation. The decoder generates response out of the fixed size vector representation. The seq2seq models were initially used for machine translation. However, we can use them as generative dialogue managers \cite{seq2seqConversation}.

Memory networks are a class of dialogue managers, which combines inference component with a long term memory. Memory network can read from and write to the memory. The model learns how to operate the memory efficiently. A memory network consists of four components, which can be trained separately \cite{memoryNetworks} or end-to-end \cite{endToEndMemoryNetworks}. 

\section{Model Architectures}
We propose two new architectures of Hybrid Code Networks. The baseline model of Hybrid Code Networks is displayed in Figure \ref{fig:SchemaOfHCN}. It uses an average of word embeddings (utterance embedding) and bag-of-words as input features.

\begin{figure}[]
\begin{center}
\includegraphics[width=\linewidth]{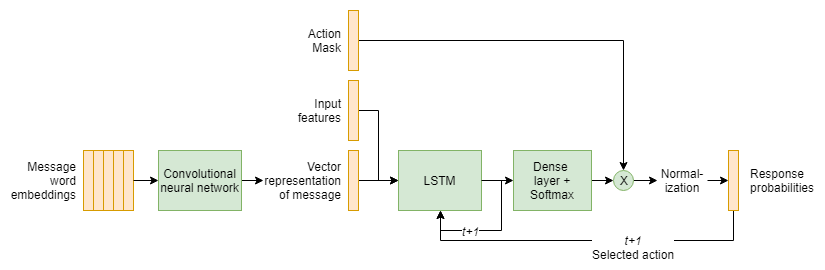}
\caption{Schema of Hybrid Code Networks with the convolutional input layer} 
\label{fig:SchemaOfHCNCNN}
\end{center}
\end{figure}

\begin{figure}[]
\begin{center}
\includegraphics[width=\linewidth]{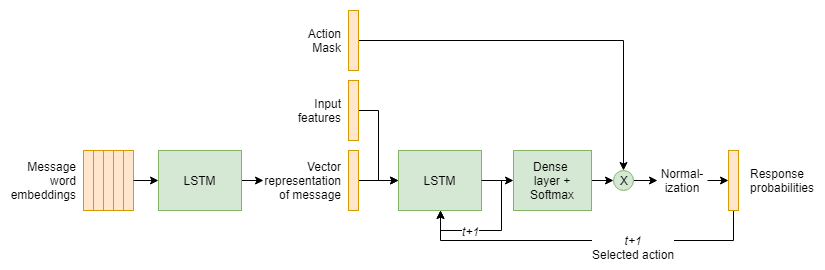}
\caption{Schema of Hybrid Code Networks with the recurrent input layer} 
\label{fig:SchemaOfHCNRNN}
\end{center}
\end{figure}

The first proposed architecture uses a convolutional neural network to create input features instead of an average of word embeddings and bag-of-words. The architecture is displayed in Figure \ref{fig:SchemaOfHCNCNN}. The architecture of the convolutional neural network is described in detail in \cite{ConvolutionalNeuralNetwork}. The only change we did to the convolutional neural network was removing the classification layer. The convolutional neural network is trained jointly with the rest of the model of the dialogue manager.
The second architecture uses a recurrent neural network to create input features. We use LSTM recurrent cell \cite{LSTM}. The architecture is displayed in Figure \ref{fig:SchemaOfHCNRNN}.  Both proposed model architectures take input message as a list of word embedding vectors.

\section{Datasets}
We tested the proposed architectures on Dialogue bAbI Task 6 \cite{bAbI} and Alquist Conversational Dataset.

The Dialogue bAbI Task 6 dataset consists of real human-bot dialogues and knowledge base from the restauration reservation domain. The dataset consists of 3,200 training dialogues, 400 validation dialogues, and 400 testing dialogues. The task is to select one of 56 possible responses.

The Alquist Conversational Dataset consists of conversations between users and the socialbot Alquist \cite{Alquist2}. It is a non-public dataset. It was collected during the 2017 Alexa Prize competition \cite{AlexaPrize2017}. The dataset consists of 37,000 dialogues about books. There are 340,000 message-response pairs in total. The average length of dialogues is 9.11 pairs, the median is 7 pairs, and there are 23,000 unique responses. The responses are clustered into 30 semantically similar classes. The goal is to select one of 30 response classes. The dataset is noisy and hard to learn because it contains voice recognition errors and part of the messages come from uncooperative users. Messages from uncooperative users are hard to interpret or out of the domain of books.

\section{Experiments}
\subsection{Experiments on Dialogue bAbI Task 6} \label{Experiments_bAbI}
We tested the proposed architectures on Dialogue bAbI task 6 two times, each time with different set of embedding vectors. We used the pretrained 300-dimensional word2vec embedding vectors trained on News\footnote{GoogleNews-vectors-negative300.bin} and 300-dimensional fastText embedding vectors pretrained on the training set of Dialogue bAbI Task 6 dataset. The fastText embeddings were trained for 100 epochs. We fixed the values of both types of embeddings during training of whole models.

We performed the hyperparameter optimization using Bayesian optimization on the validation part of the dataset. We used the implementation from Scikit-Optimize library\footnote{https://scikit-optimize.github.io/}. We performed 30 rounds of training with a different set of hyperparameters for each model. Each training was 30 epochs long. We evaluated the performance of the model after each epoch and saved the weights of the model achieving the highest turn accuracy on the validation set for each set of hyperparameters. We selected the hyperparameters achieving the highest turn accuracy. The values of selected hyperparameters are presented in table \ref{hyperparameters}.

\begin{table*}[h]
\resizebox{\textwidth}{!}{%
\begin{tabular}{l|l|l|l|l|l|l}
\hline
\textbf{Parameter/Model}           & \textbf{fastText} & \textbf{fastText+CNN} & \textbf{fastText+RNN} & \textbf{word2vec} & \textbf{word2vec+CNN} & \textbf{word2vec+RNN} \\ \hline
LSTM size                 & 55        & 245            & 505            & 85        & 109            & 219            \\
Input LSTM size           & -        & -            & 199            & -        & -            & 312            \\
Convolutional filters           & -        & 21            & -            & -        & 6            & -            \\
LSTM dropout              & 0.85        & 0.80            & 0.94            & 0.92        & 0.79            & 0.74            \\
Input LSTM dropout        & -        & -            & 0.97            & -        & -            & 0.91            \\
Convolutional dropout     & -        & 0.72            & 0            & -        & 0.84            & -            \\
Fully connected dropout   & 0.82        & 0.79            & 0.76            & 0.59        & 0.93            & 0.98            \\
Learning rate             & 0.008        & 0.0001            & 0.0003            & 0.001        & 0.005            & 0.00005            \\
Activation function       & relu        & relu            & relu            & tanh        & tanh            & relu            \\
Input activation function & -        & -            & tanh            & -        & -            & tanh            \\
Adam epsilon              & 1E-8        & 1E-8            & 1E-8            & 1E-8        & 0.1            & 1E-8            \\
Adam beta1                & 0.9        & 0.5            & 0.5            & 0.5        & 0.5            & 0.9            \\ \hline
Turn accuracy             & 69.4\%        & 71.5\%            & 68.0\%            & 71.3\%        & 70.4\%            & 65.5\%            \\ \hline
\end{tabular}}
\caption{Model's hyperparameters achieving the highest turn accuracy on the validation set of Dialogue bAbI Task 6}
\label{hyperparameters}
\end{table*}

We trained each model with the hyperparameters, which achieved the highest validation accuracy. We used 12 training epochs. We measured the turn accuracy on validation set after each training epoch and saved the weights achieving the highest validation turn accuracy. We used these weights for testing. Results are presented in section \ref{Results}.

\subsection{Experiments on Alquist Conversational Dataset}
We measured the performance of all models on the Alquist Conversational Dataset. We used the word2vec embedding vectors pretrained on News and fastText embedding vectors trained for 100 epochs on the training set of Alquist Conversational Dataset. We did not perform hyperparameter optimization due to a long time of training. We used hyperparameters presented in section \ref{Experiments_bAbI} for each model instead.

We trained models for 12 epochs. We measured the turn accuracy after each epoch and saved the weights which achieved the highest turn accuracy. We measured turn and dialogue accuracies on the testing set using saved weights. Results are presented in section \ref{Results}.

\section{Results} \label{Results}
We present the accuracy of described models in table \ref{testingAccuracy}. Baseline Hybrid Code Networks outperforms other models on the Dialogue bAbI task 6 in both turn and dialogue accuracy \cite{HybridCodeNetworks}. The Hybrid Code Networks using a convolutional neural network as input layer and fastText embeddings improves the turn accuracy of the baseline Hybrid Code Networks by 3.3\% on the Dialogue bAbI task 6 dataset. The dialogue accuracy of this model is smaller by 1.4\%. However, we were not able to reproduce the dialogue accuracy described in \cite{HybridCodeNetworks} with our baseline model (labeled as word2vec).

Hybrid Code Network model achieves the highest turn accuracy of 92.6\% on the Alquist dataset if the model uses a convolutional neural network as an input layer and word2vec embeddings. This model achieves the second highest dialogue accuracy.

The results show that convolutional neural network as an input layer in Hybrid Code Networks achieves higher turn accuracy than baseline Hybrid Code Networks and Hybrid Code Network using a recurrent neural network as an input layer.

\begin{table*}[t]
\centering
\resizebox{\textwidth}{!}{%
\begin{tabular}{c|cc|cc}
\hline
                                 & \multicolumn{2}{c|}{\textbf{bAbI6}} & \multicolumn{2}{c}{\textbf{Alquist}} \\
\textbf{Model}                   & Turn Acc.         & Dialogue Acc.   & Turn Acc.       & Dialogue Acc.      \\ \hline
Bordes and Weston (2017) \cite{LearningEndToEndGoalOrientedDialog}         & 41.1\%            & 0.0\%           & -               & -                  \\
Liu and Perez (2016) \cite{GatedEndToEndMemoryNetworks}             & 48.7\%            & 1.4\%           & -               & -                  \\
Eric and Manning (2017)  \cite{CopyAugmentedSequenceToSequence}        & 48.0\%            & 1.5\%           & -               & -                  \\
Seo et al. (2016)  \cite{QueryRegressionNetworksForMachineComprehension}             & 51.1\%            & -               & -               & -                  \\
Williams, Asadi and Zweig (2017) \cite{HybridCodeNetworks} & 55.6\%            & \textbf{1.9\%}  & -               & -                  \\ \hline
fastText                         & 57.6\%            & 0.8\%           & 86.9\%          & 51.7\%             \\
fastText+CNN                     & \textbf{58.9\%}   & 0.5\%           & 90.6\%          & 63.0\%             \\
fastText+RNN                     & 54.9\%            & 0.3\%           & 80.6\%          & 40.5\%             \\
word2vec                         & 57.4\%            & 0.4\%           & 92.2\%          & \textbf{68.0\%}    \\
word2vec+CNN                     & 56.3\%            & 0.1\%           & \textbf{92.6\%} & 67.8\%             \\
word2vec+RNN                     & 54.6\%            & 0.1\%           & 83.9\%          & 45.2\%             \\ \hline
\end{tabular}%
}
\caption{Testing accuracy of dialogue managers on Dialogue bAbI task 6 and Alquist Conversational Dataset}
\label{testingAccuracy}
\end{table*}

\section{Conclusion}
We proposed two new architectures of Hybrid Code Networks. They use convolutional and recurrent neural networks as input layers, instead of the average of word embeddings and bag-of-words. We tested the turn and dialogue accuracy of the proposed architectures on the Dialogue bAbI task 6 and Alquist Conversational Dataset.

Results show that Hybrid Code Networks using a convolutional neural network as input layer improves the turn accuracy. The model using convolutional neural network outperforms the baseline model of Hybrid Code Networks on the Dialogue bAbI task 6 and Alquist Conversational Dataset.

\section*{Acknowledgements}
The research described in the paper was supervised by Ing. J. \v{S}ediv\'{y}, CSc. CIIRC CTU in Prague and Ing. V\'{a}clav Chud\'{a}\v{c}ek, Ph.D. CIIRC CTU in Prague, and supported by the Grant Agency of the Czech Technical University in Prague, grant No. SGS19/091/OHK3/1T/37.


\begin{authorcv}{Petr MAREK}
is a Ph.D. student of conversational artificial intelligence at Faculty of Electrical Engineering, CTU. He works on conversational AI Alquist, which was the second prize winner of Amazon Alexa Prize 2017 and 2018. He finished his master degree in 2018 in artificial intelligence at FEE CTU in 2018. The topic of his master thesis was Dialog manager for conversational AI.
\end{authorcv}

\end{document}